\documentstyle{llncs}
\begin{document}
\title{On Sequences with Non-Learnable Subsequences}

\author{Vladimir V. V'yugin}
\institute{Institute for Information Transmission Problems,
Russian Academy of Sciences,
Bol'shoi Karetnyi per. 19, Moscow GSP-4, 127994, Russia\\
e:mail vyugin@iitp.ru}

\maketitle
\begin{abstract}
The remarkable results of Foster and Vohra was a starting point
for a series of papers which show that
any sequence of outcomes can be learned (with no prior knowledge)
using some universal randomized
forecasting algorithm and forecast-dependent checking rules.
We show that for the class of all
computationally efficient outcome-forecast-based checking rules,
this property is violated.
Moreover, we present a probabilistic algorithm generating with probability
close to one a sequence with a subsequence which
simultaneously miscalibrates all partially weakly computable randomized
forecasting algorithms.

According to the Dawid's prequential framework we consider
partial recursive randomized algorithms.
\end{abstract}

\section{Introduction}\label{intr-1}

Let a binary sequence $\omega_1,\omega_2,\dots ,\omega_{n-1}$ of outcomes
is observed by a forecaster whose task is to give a probability $p_n$
of a future event $\omega_n=1$.
The evaluation of probability forecasts is based on a method called
{\it calibration}: informally, following Dawid~\cite{Daw82} forecaster is
said to be well-calibrated if for any $p^*$ the event $\omega_n=1$ holds
in $100p^*\%$ of moments of time as he choose $p_n\approx p^*$.
(see also~\cite{Daw85}).

Let us give some notations. Let $\Omega$ be the set of all
infinite binary sequences, $\Xi$ be the set of all finite binary
sequences and $\lambda$ be the empty sequence. For any finite or
an infinite sequence $\omega=\omega_{1}\ldots\omega_{n}\ldots$, we
write $\omega^n=\omega_{1}\dots\omega_{n}$ (we put
$\omega_0=\omega^0=\lambda$). Also, $l(\omega^n)=n$ denotes the
length of the sequence $\omega^n$. If $x$ is a finite sequence and
$\omega$ is a finite or infinite sequence then $x\omega$ denotes
the concatenation of these sequences, $x\sqsubseteq\omega$ means
that $x=\omega^n$ for some $n$.

In the measure-theoretic framework we expect that the forecaster has a method
for assigning probabilities $p_n$ of a future event $\omega_n=1$ for all
possible finite sequences $\omega_1,\omega_2,\dots ,\omega_{n-1}$.
In other words, all conditional probabilities
$$
p_n=P(\omega_n=1|\omega_1,\omega_2,\dots ,\omega_{n-1})
$$
must be specified and the overall probability distribution in the space
$\Omega$ of all infinite binary sequences will be defined.
But in reality, we should recognize that we have only individual
sequence $\omega_1,\omega_2,\dots ,\omega_{n-1}$ of events and that
the corresponding forecasts $p_n$ whose testing is considered may fall
short of defining a full probability distribution in the whole space $\Omega$.
This is the point of {\it the prequential principle} proposed by
Dawid~\cite{Daw82}. This principle says that the evaluation
of a probability forecaster should depend only on his actual probability
forecasts and the corresponding outcomes. The additional information
contained in a probability measure that has these probability forecasts
as conditional probabilities should not enter in the evaluation.
According to Dawid's prequential framework we do not consider
numbers $p_n$ as conditional probabilities generated by some
overall probability distribution defined for all possible events.
In such a way, {\it a deterministic forecasting system} is
{\it a partial recursive} function $f:\Xi\to [0,1]$.
We suppose that a valid forecasting system $f$ is defined
on all finite initial fragments $\omega_1,\dots ,\omega_{n-1},\dots$ of
an analyzed individual sequence of outcomes.

First examples of individual sequences for which well-calibrated
deterministic forecasting is impossible (non-calibrable sequences)
were presented by Oakes~\cite{Oak85} (see also Shervish~\cite{Sch85}).
Unfortunately, the methods used in these
papers, and in Dawid~\cite{Daw82},~\cite{Daw85}, do not comply
with prequential principle; they depend on some mild assumptions
about the measure from which probability forecasts are derived as
conditional probabilities. The method of generation the
non-calibrable sequences with probability arbitrary close to one
presented in V'yugin~\cite{Vyu98} also is based on the same
assumptions. In this paper we modify construction
from~\cite{Vyu98} for the case of partial deterministic and
randomized forecasting systems
do not corresponding to any overall probability distributions.

Oakes~\cite{Oak85} showed that any everywhere defined forecasting
system $f$ is not calibrated for a sequence $\omega=\omega_1\omega_2\dots$
defined
\[
\omega_i=
  \left\{
    \begin{array}{l}
      1 \mbox{ if } p_i<0.5
    \\
      0 \mbox{ otherwise }
    \end{array}
  \right.
\]
and $p_i=f(\omega_1\dots\omega_{i-1})$, $i=1,2,\dots$.

Foster and Vohra~\cite{FoV98} showed that the well-calibrated
forecasts are possible if these forecasts are randomized.
By {\it a randomized}~ forecasting system they mean a random
variable $f(\alpha;x)$ defined on some probability space
$\Omega_x$ supplied by some probability distribution $Pr_x$, where
$x\in\Xi$ is a parameter. As usual, we omit the argument $\alpha$.
For any infinite $\omega$, these probability distributions
$Pr_{\omega^{i-1}}$ generate the {\it overall probability
distribution} $Pr$ on the direct product of probability spaces
$\Omega_{\omega^{i-1}}$, $i=1,2,\dots$.

It was shown in~\cite{FoV98},~\cite{KaF2004} that any sequence
can be learned:
for any $\Delta>0$, a universal randomized forecasting system $f$ was
constructed such that for any sequence $\omega=\omega_1\omega_2\dots$
the overall probability $Pr$ of the event
\begin{eqnarray}\label{calib-1}
\left|\frac{1}{n}\sum_{i=1}^{n}I(\tilde p_i)(\omega_i-\tilde p_i)\right|\le
\Delta
\end{eqnarray}
tends to one as $n\to\infty$, where $\tilde p_i=f(\omega^{n-1})$ is the random
variable, $I(p)$ is the characteristic function of an arbitrary subinterval
of $[0,1]$; we call this function a {\it forecast-based} checking rule.

Lehrer~\cite{Lehr97} and Sandrony et al.~\cite{sansmovo03} extended
the class of checking rules to combination of {\it forecast-} and
{\it outcome-based } checking rules: a checking rule is a function
$c(\omega^{i-1},p)=\delta(\omega^{i-1})I(p)$, where $\delta:\Xi\to\{0,1\}$
is an outcome-based checking rule, and $I(p)$ is a characteristic function
of a subinterval of $[0,1]$. They also considered a more general class of
randomized forecasting systems - random variables
$\tilde p_i=f(\alpha;\omega^{i-1},p^{i-1})$, where
$p^{i-1}=p_1,\dots ,p_{i-1}$ is the
sequence of past realized forecasts.

For $k=1,2,\dots$, let $\{\delta_k\}$ be any sequence of outcome-based
checkng rules and $\{I_k\}$ be any sequence of characteristic functions
of subintervals of $[0,1]$. Sandrony et al.~\cite{sansmovo03} defined
a randomized universal forecasting system which calibrates all checking rules
$\{\delta_k I_k\}$, $k=1,2,\dots$, i.e., such that for any $\Delta>0$ and
for any sequence $\omega=\omega_1\omega_2\dots$,
the overall probability of the event (\ref{calib-1}) tends to one as
$n\to\infty$, where $\tilde p_i=f(\omega^{n-1},p^{i-1})$ and $I(\tilde p_i)$
is replaced on $\delta_k(\omega^{i-1})I_k(\tilde p_i)$ for all $k=1,2,\dots$.

In this paper we consider the class of all computable (partial
recursive) outcome-based checking rules $\{\delta_k\}$ and a
slightly different class of randomized forecasting systems: our
forecasting systems are random variables $\tilde
p_i=f(\alpha;\omega^{i-1})$ do not depending on past realized
forecasts (this take a place for the universal forecasting systems
defined in~\cite{FoV98} and~\cite{Vov2007} \footnote { Note that
the algorithm from~\cite{sansmovo03} can be modified in a fashion
of~\cite{FoV98}, i.e., such that at any step of the construction
past forecasts can be replaced on measures with finite supports
defined on previous steps. Since these measures are defined
recursively in the process of the construction, they can be
eliminated from the condition of the universal forecasting
algorithm. } ). Concurrently, such a function can be undefined
outside $\omega$, it requires that any well defined forecasting
system must be defined on all initial fragments of an analyzed
sequence of outcomes. This peculiarity is important, since we
consider forecasting systems possessing some computational
properties: there is an algorithm computing the probability
distribution function of such forecasting system. This algorithm
when fed to some input can never finish its work, and so, is
undefined on this input.

In this case, a universal randomized
forecasting algorithm which calibrates all computationally efficient
outcome-forecast-based checking
rules does not exist. Moreover, we construct {\it a probabilistic
generator} (or probabilistic algorithm) of {\it non-learnable} (in
this way) sequences. This generator outputs with probability close
to one an infinite sequence such that for each randomized
forecasting system $\tilde p_i=f(\alpha;\omega^{i-1})$ some
computable outcome-based checking rule $\delta$ selects an infinite
subsequence of $\omega$ on which the property (\ref{calib-1})
fails for some characteristic function $I$ with the overall
probability one, where the overall probability is associated with
the forecasting system $f$.

\section{Miscalibrating the forecasts} \label{sec-net-1}

We use standard notions of the theory of algorithms. This theory is
systematically treated in, for example, Rogers \cite{Rog67}.
We fix some effective one-to-one enumeration
of all pairs (triples, and so on) of nonnegative integer numbers.
We identify any pair $(t,s)$ and its number $\langle t,s\rangle$;
let $p(\langle t,s\rangle)=t$.

A function $\phi\colon A\rightarrow \cal R$ is called (lower) semicomputable if
$\{(r,x):r<\phi(x)\}$ ($r$ is a rational number) is a recursively
enumerable set. A function $\phi$ is upper semicomputable if $-\phi$ is
lower semicomputable.
Standard argument based on the recursion theory shows that
there exist the lower and upper semicomputable real functions
$\phi^{-}(j,x)$ and $\phi^{+}(k,x)$ universal
for all lower semicomputable and upper semicomputable functions
from $x\in\Xi$; in particular every computable real
function $\phi(x)$ can be represented as
$\phi(x)=\phi^{-}(j,x)=\phi^{+}(k,x)$
for all $x$, for some $j$ and $k$.
Let $\phi_{s}^-(j,x)$ be equal to the maximal rational number $r$ such that
the triple $(r,j,x)$ is enumerated in $s$ steps in the process of
enumerating of the set
$\{(r,j,x):r<\phi(j,x),\mbox{ }r \mbox{ is rational} \}$
and equals $-\infty$, otherwise.
Any such function $\phi^-_s(j,x)$ takes only finite number of rational
values distinct from $-\infty$.
By definition, $\phi_{s}^-(j,x)\le \phi_{s+1}^-(j,x)$ for all $j,s,x$,
and $\phi^-(j,x)=\lim\limits_{s\to\infty}\phi_{s}^-(j,x).$
An analogous non-increasing sequence of functions $\phi_{s}^+(k,x)$
exists for any upper semicomputable function.

Let $i=\langle t,k\rangle$.
We say that a real function $\phi_i(x)$ is {\it defined on} $x$ if given any
degree of precision - positive rational number $\kappa>0$, it holds
$|\phi_{s}^+(t,x)-\phi_{s}^-(k,x)|\le\kappa$
for some $s$; $\phi_i(x)$
undefined, otherwise. If any such $s$ exists then for minimal such $s$,
$\phi_{i,\kappa}(x)=\phi_{s}^-(k,x)$ is called the rational approximation
(from below) of $\phi_i(x)$ up to $\kappa$; $\phi_{i,\kappa}(x)$ undefined,
otherwise.

To define a measure $P$ on $\Omega$, we define values $P(z)=P(\Gamma_z)$
for all intervals $\Gamma_z=\{\omega\in\Omega:z\sqsubseteq\omega\}$,
where $z\in\Xi$, and extend this function on all Borel subsets of
$\Omega$ in a standard way.

We use also a concept of {\it computable operation} on $\Xi\bigcup\Omega$
(see \cite{ZvL70}). Let $\hat F$ be a recursively enumerable
set of ordered pairs of finite sequences satisfying the following properties:
(i) $(x,\lambda)\in\hat F$ for each $x$;
(ii) if $(x,y)\in\hat F$, $(x',y')\in\hat F$ and $x\sqsubseteq x'$ then
$y\sqsubseteq y'$ or $y'\sqsubseteq y$ for all finite binary sequences
$x,x',y,y'$.
A computable operation $F$ is defined as follows
$$
F(\omega)=\sup\{y\mid x\sqsubseteq\omega\mbox{ and }
(x,y)\in\hat F\mbox{ for some } x\},
$$
where $\omega\in\Omega\bigcup\Xi$ and $\sup$ is in the sense of the
partial order $\sqsubseteq$ on $\Xi$.

{\it A probabilistic algorithm} is a pair $(L,F)$, where
$L(x)=L(\Gamma_x)=2^{-l(x)}$ is the uniform measure on $\Omega$ and
$F$ is a computable operation. For any probabilistic algorithm $(L,F)$ and
a set $A\subseteq\Omega$, we consider the probability
$L\{\omega:F(\omega)\in A\}$
of generating by means of $F$ a sequence from $A$ given
a uniformly distributed sequence $\omega$.

A partial randomized forecasting system $f$ is
{\it weakly computable} if its {\it weak probability distribution
function}
$\varphi_n(\omega^{n-1})=Pr_n\{f(\omega^{n-1})<\frac{1}{2}\}$
is a partial recursive function from $\omega^{n-1}$.

Any function $\delta:\Xi\to\{0,1\}$ is called an outcome-based selection
(or checking) rule.
For any sequence $\omega=\omega_1\omega_2\dots$, the selection rule $\delta$
selects a sequence of indices $n_i$ such that
$\delta(\omega^{n_i-1})=1$, $i=1,2,\dots$,
and the corresponding subsequence $\omega_{n_1}\omega_{n_2}\dots$ of $\omega$.

The following theorem is the main result of this paper.
In particular, it shows that the construction of the universal forecasting
algorithm from Sandrony et al.~\cite{sansmovo03} is
computationally non-efficient in a case when the class of all partial
recursive outcome-based checking rules $\{\delta_k\}$ is used.
\begin{theorem}\label{main-1}
For any $\epsilon>0$ a probabilistic algorithm $(L,F)$ can be constructed,
which with probability $\ge 1-\epsilon$ outputs
an infinite binary sequence $\omega=\omega_1\omega_2\dots$ such that for every
partial weakly computable randomized forecasting system $f$
defined on all initial fragments of the sequence $\omega$ there
exists a computable selection rule $\delta$ defined on all these
fragments and such that for $\nu=0$ or for $\nu=1$ the overall probability
of the event
\begin{eqnarray}\label{main-1_2}
\limsup_{n\to\infty}\left|\frac{1}{n}\sum_{i=1}^{n}
\delta(\omega^{i-1})I_\nu(\tilde p_i)(\omega_i-\tilde p_i)\right|\ge 1/16
\end{eqnarray}
equals one, where $I_0$ and $I_1$ are the characteristic functions of
the intervals $[0,\frac{1}{2})$ and $[\frac{1}{2},1]$,
$\tilde p_i=f(\omega^{i-1})$ is a random variable,
$i=1,2,\dots$, and the overall probability distribution is
associated with $f$.
\end{theorem}
{\it Proof}.
For any probabilistic algorithm $(L,F)$, we consider the function
\begin{eqnarray}\label{five-1}
Q(x)=L\{\omega:x\sqsubseteq F(\omega)\}.
\end{eqnarray}
It is easy to verify that this function is lower semicomputable and
satisfies: $Q(\lambda)\le 1$; $Q(x0)+Q(x1)\leq Q(x)$ for all $x$.
Any function satisfying these properties is called semicomputable
semimeasure. For any semicomputable semimeasure
$Q$ a probabilistic algorithm $(L,F)$ exists such that (\ref{five-1})
holds.
Though the semimeasure $Q$ is not a measure, we consider the corresponding
measure on the set $\Omega$
$$
\bar Q(\Gamma_x)=\inf\limits_n\sum\limits_{l(y)=n,x\sqsubseteq y}Q(y).
$$
We will construct a semicomputable semimeasure $Q$ as a some sort of network
flow. We define an infinite network on the base of the infinite binary tree.
Any $x\in\Xi$ defines two edges $(x,x0)$ and $(x,x1)$ of length one.
In the construction below we will mount to the network extra edges $(x,y)$
of length $>1$, where $x,y\in\Xi$, $x\sqsubseteq y$ and $y\not =x0, x1$.
By the length of the edge $(x,y)$ we mean the number $l(y)-l(x)$.
For any edge $\sigma=(x,y)$ we denote by $\sigma_1=x$ its starting
vertex and by $\sigma_2=y$ its terminal vertex.
A computable function $q(\sigma)$ defined on all edges of length one and
on all extra edges and taking rational values is called {\it a network}
if for all $x\in\Xi$
\begin{eqnarray*} \label{net-1}
\sum\limits_{\sigma:\sigma_1=x} q(\sigma)\le 1.
\end{eqnarray*}
Let $G$ be the set of all extra edges of the network $q$
(it is a part of the domain of $q$).
By $q$-{\it flow } we mean the minimal semimeasure $P$ such that
$P\ge R$, where the function $R$ is defined by the following
recursive equations $R(\lambda)=1$ and
\begin{eqnarray}
R(y)=\sum\limits_{\sigma:\sigma_2=y}q(\sigma)R(\sigma_1)
\label{net-base-2}
\end{eqnarray}
for $y\not =\lambda$.
A network $q$ is called {\it elementary} if the set of extra edges
is finite and $q(\sigma)=1/2$ for almost all edges of unit length.
For any network $q$, we define the {\it network flow delay} function
($q$-delay function)
\begin{eqnarray*}
d(x)=1-q(x,x0)-q(x,x1).
\end{eqnarray*}
The construction below works with all computable real functions
$\phi_t(x)$, $x\in\Xi$, $t=1,2,\dots$. We suppose that for any computable
function $\phi$ there exist infinitely many programs $t$ such that
$\phi_t=\phi$.
\footnote
{
To obtain this property, we can replace the sequence $\phi_t(x)$
on a sequence $\phi'_{\langle t,s\rangle}(x)=\phi_t(x)$ for all $s$.
}
Any pair $i={\langle t,s\rangle}$ is considered as a program
for computing the rational approximation $\phi_{t,\kappa_s}(\omega^{n-1})$
of $\phi_t$ from below up to $\kappa_s=1/s$.

By the construction below we visit any function $\phi_t$ on infinitely
many steps $n$.
To do this, we use the function $p(n)$: for any positive integer number
$i$ we have $p(n)=i$ for infinitely many $n$.

Let $\beta$ be a finite sequence and $1\le k<l(\beta)$. A bit $\beta_k$
of the sequence $\beta$ is called {\it hardly predictable} by a
program $i=\langle t,s\rangle$ if $\phi_{t,\kappa_s}(\beta^{k-1})$ is
defined and
\[
  \beta_k=
  \left\{
    \begin{array}{l}
      0 \mbox{ if } \phi_{t,\kappa_s}(\beta^{k-1})\ge \frac{1}{2}
    \\
      1 \mbox{ otherwise }
    \end{array}
  \right.
\]
\begin{lemma}\label{num-hard}
Let $i=\langle t,s\rangle$ be a program and $\mu$ be an arbitrary
sufficiently small positive real number.
Then for any binary sequence $x$ of length $n$
the portion of all sequences $\gamma$ of length
$K=\lceil (2+\mu)i\rceil n$
(in the set of all finite sequences of length $K$) such that

1) $\phi_{t,\kappa_s}(x\gamma^k)$ is defined for all $0\le k<K$,

2) the number of hardly predictable bits of $\gamma$ by the
forecasting program $i$ is less than $in$,
\\
is $\le 2^{-2\mu^2 in+O(\log (in))}$ for all sufficiently large $n$.
\end{lemma}
{\it Proof}. Any function $\sigma(x)$, where $x\in\Xi$ and
$\sigma(x)\in\{A,B\}$, is called {\it labelling} if
$\sigma(x0)\not =\sigma(x1)$ for all $x\in\Xi$.
For any $\gamma$ of length $K$ and for any $k$ such that $1\le k<K$,
define $\sigma(\gamma^{k+1})=A$ and $\sigma(\gamma^k\bar\gamma_{k+1})=B$ if the
bit $\gamma_{k+1}$ of the sequence $x\gamma$ is hardly predictable,
where we denote $\bar\theta=1-\theta$ for any binary bit $\theta$.
Since $\phi_{t,\kappa_s}(x\gamma^k)$ is defined for all
$0\le k<K$, then $\sigma(\gamma^{k+1})$ is also defined for all these $k$.
This partial labelling $\sigma$ can be easily extended on the
set of all binary sequences of length $K$ in many different ways.
We fix some such extension.
Then the total number of all $\gamma$ satisfying 1)-2) does not exceed
the total number of all binary sequences of length $K$ with
$\le in$ labels $A$. Therefore, for all sufficiently large $n$,
the portion of these $\gamma$ does not exceed
$$
\sum\limits_{i\le in}{K\choose i}2^{-K}
\le 2^{-(1-H(1/2-\mu))2in+O(\log(in))}
\le 2^{-2\mu^2in+O(\log(in))},
$$
where $H(r)=-r\log r-(1-r)\log(1-r)$.
$\Box$

In the following we put $\mu=1/\log (i+1)$.

We define an auxiliary relation $B(i,q^{n-1},\sigma,n)$ and a
function $\beta(x,q^{n-1},n)$. Let $x,\beta\in\Xi$. The value of
$B(i,q^{n-1},(x,\beta),n)$ is {\it true} if the following
conditions hold:
\begin{itemize}
\item{}
$n\ge (1+\lceil (2+\log^{-1}(i+1))i\rceil)l(x)$;
\item{}
$l(\beta)=n$ and $x\sqsubseteq\beta$;
\item{}
$d^{n-1}(\beta^j)<1$ for all $j$ such that $1\le j<n$;
\item{}
for all $j$, $l(x)<j\le (1+\lceil (2+\log^{-1}(i+1))i\rceil) l(x)$, the value
$\phi_{t,\kappa_s}(\beta^{j-1})$
is computed in $\le n$ steps, and for at least $il(x)$ of these $j$
the bit $\beta_j$ is hardly predictable by the program $i=\langle t,s\rangle$.
\end{itemize}
The value of $B(i,q^{n-1},(x,\beta),n)$ is {\it false}, otherwise.
Define
\begin{eqnarray*}
\beta(x,q^{n-1},n)=\min\{y:p(l(y))=p(l(x)),B(p(l(x)),q^{n-1},(x,y),n)\}.
\end{eqnarray*}
Here $\min$ is considered for lexicographical ordering of strings; we suppose
that $\min\emptyset$ is undefined.

{\bf Construction.}
Let $\rho(n)=(n+n_0)^2$ for some sufficiently large $n_0$ (the value $n_0$
will be specified below in the proof of Lemma~\ref{nontriv-1a}).

Using the mathematical induction by $n$, we define a sequence $q^n$
of elementary networks. Put $q^0(\sigma)=1/2$ for all edges $\sigma$
of length one.

Let $n>0$ and a network $q^{n-1}$ is defined.
Let $d^{n-1}$ be the $q^{n-1}$-delay function
and let $G^{n-1}$ be the set of all extra edges. We suppose also that
$l(\sigma_2)<n$ for all $\sigma\in G^{n-1}$.

Let us define a network $q^n$. At first, we define
a network flow delay function $d^n$ and a set $G^n$.
The construction can be split up into two cases.

Let $w(i,q^{n-1})$ be equal to the minimal
$m$ such that $p(m)=i$ and $m>l(\sigma_2)$ for each extra edge
$\sigma\in G^{n-1}$ such that $p(l(\sigma_1)))<i$.

The inequality $w(i,q^m)\not =w(i,q^{m-1})$
can be induced by some task $j<i$ that mounts an extra edge $\sigma=(x,y)$
such that $l(x)>w(i,q^{m-1})$ and $p(l(x))=p(l(y))=j$.
Lemma~\ref{gen-tech-1} (below) will show that this can happen only at finitely
many steps of the construction.

{\it Case 1}. $w(p(n),q^{n-1})=n$ (the goal of this part is
to start a new task $i=p(n)$ or to restart the existing task $i=p(n)$ if it
was destroyed by some task $j<i$ at some preceding step).

Put $d^n(y)=1/\rho(n)$ for $l(y)=n$ and define
$d^n(y)=d^{n-1}(y)$ for all other $y$. Put also $G^n=G^{n-1}$.

{\it Case 2.} $w(p(n),q^{n-1})<n$ (the goal of this part is to process
the task $i=p(n)$).
Let $C_n$ be the set of all $x$ such that $w(i,q^{n-1})\le l(x)<n$,
$0<d^{n-1}(x)<1$, the function $\beta(x,q^{n-1},n)$ is defined
\footnote
{
In particular, $p(l(x))=i$ and $l(\beta(x,q^{n-1},n))=n$.
}
and there is no extra edge $\sigma\in G^{n-1}$ such that $\sigma_1=x$.

In this case for each $x\in C_n$ define $d^n(\beta(x,q^{n-1},n))=0$,
and for all other $y$ of length $n$ such that $x\sqsubset y$ define
$$
d^n(y)=\frac{d^{n-1}(x)}{1-d^{n-1}(x)}.
$$
Define $d^n(y)=d^{n-1}(y)$ for all other $y$.
We add an extra edge to $G^{n-1}$, namely, define
\begin{eqnarray*}
G^n=G^{n-1}\cup\{(x,\beta(x,q^{n-1},n)):x\in C_n\}.
\end{eqnarray*}
We say that the task $i=p(n)$ {\it mounts} the extra edge
$(x,\beta(x,q^{n-1},n))$ to the network
and that all existing tasks $j>i$ are destroyed by the task $i$.

After Case 1 and Case 2, define for any edge $\sigma$ of unit length
$$
q^n(\sigma)=\frac{1}{2}(1-d^n(\sigma_1))
$$
and $q^n(\sigma)=d^n(\sigma_1)$ for each extra edge $\sigma\in G^n$.

{\it Case 3}. Cases 1 and 2 do not hold.
Define $d^n=d^{n-1}$, $q^n=q^{n-1}$, $G^n=G^{n-1}$.

As the result of the construction we define the network
$q=\lim\limits_{n\to\infty}q^n$, the network flow delay function
$d=\lim\limits_{n\to\infty}d^n$ and the set of extra edges $G=\cup_n G^n$.

The functions $q$ and $d$ are computable and the set $G$ is recursive
by their definitions. Let $Q$ denotes the $q$-flow.

The following lemma shows that any task can mount new extra edges only
at finite number of steps.
Let $G(i)$ be the set of all extra edges mounted by the task $i$,
$w(i,q)=\lim_{n\to\infty} w(i,q^n)$.
\begin{lemma} \label{gen-tech-1}
The set $G(i)$ is finite, $w(i,q)$ exists and $w(i,q)<\infty$ for all $i$.
\end{lemma}
{\it Proof.} Note that if $G(j)$ is finite for all $j<i$,
then $w(i,q)<\infty$. Hence, we must prove that the set
$G(i)$ is finite for any $i$. Suppose that the opposite assertion holds.
Let $i$ be the minimal such that $G(i)$ is infinite.
By choice of $i$ the sets $G(j)$ for all $j<i$ are finite.
Then $w(i,q)<\infty$.

For any $x$ such that $l(x)\ge w(i,q)$, consider
the maximal $m$ such that for some initial fragment $x^m\sqsubseteq x$
there exists an extra edge $\sigma=(x^m,y)\in G(i)$.
If no such extra edge exists define $m=w(i,q)$. By definition,
if $d(x^m)\not =0$ then $1/d(x^m)$ is an integer number. Define
\[
  u(x)=
  \left\{
    \begin{array}{l}
      1/d(x^m) \mbox{ if } d(x^m)\not = 0, l(x)\ge w(i,q)
    \\
      \rho(w(i,q)) \mbox{ if } l(x)<w(i,q)
    \\
     0 \mbox{ otherwise }
    \end{array}
  \right.
\]
By construction the integer valued function $u(x)$ has the property:
$u(x)\ge u(y)$ if $x\sqsubseteq y$. Besides, if $u(x)>u(y)$ then $u(x)>u(z)$
for all $z$ such that $x\sqsubseteq z$ and $l(z)=l(y)$. Then the function
$$
\hat u(\omega)=\min\{n:u(\omega^i)=u(\omega^n) {\rm \ for \ all \ } i\ge n\}
$$
is defined for all $\omega\in\Omega$. It is easy to see that this
function is continuous. Since $\Omega$ is compact space in the
topology generated by intervals $\Gamma_x$, this function is
bounded by some number $m$. Then $u(x)=u(x^m)$ for all $l(x)\ge
m$. By the construction, if any extra edge of $i$th type was
mounted to $G(i)$ at some step then $u(y)<u(x)$ holds for some new
pair $(x,y)$ such that $x\sqsubseteq y$. This is contradiction
with the existence of the number $m$. $\Box$

An infinite sequence $\alpha\in\Omega$ is called an $i$-{\it extension}
of a finite sequence $x$ if $x\sqsubseteq\alpha$ and
$B(i,q^{n-1},x,\alpha^n,n)$ is true for almost all $n$.

A sequence $\alpha\in\Omega$ is called $i$-{\it closed} if
$d(\alpha^n)=1$ for some $n$ such that $p(n)=i$, where $d$ is the
$q$-delay function. Note that if $\sigma\in G(i)$ is some extra
edge (i.e. an edge of $i$th type) then $B(i,q^{n-1},\sigma,n)$ is
true, where $n=l(\sigma_2)$.

\begin{lemma} \label{exten-1}
Let for any initial fragment $\omega^n$ of an infinite sequence
$\omega$ some $i$-extension exists. Then either the sequence
$\omega$ will be $i$-closed in the process of the construction or
$\omega$ contains an extra edge of $i$th type (i.e.
$\sigma_2\sqsubseteq\omega$ for some $\sigma\in G(i)$).
\end{lemma}
{\it Proof.} Let a sequence $\omega$ is not $i$-closed. By
Lemma~\ref{gen-tech-1} the maximal $m$ exists such that $p(m)=i$
and $d(\omega^m)>0$. Since the sequence $\omega^m$ has an
$i$-extension and $d(\omega^m)<1$, by Case 2 of the construction a
new extra edge $(\omega^m,y)$ of $i$th type must be mounted to the
binary tree. By the construction $d(y)=0$ and $d(z)\not =0$ for all
$z$ such that $\omega^m\sqsubseteq z$, $l(z)=l(y)$, and $z\not
=y$. By the choice of $m$ we have $y\sqsubseteq\omega$. $\Box$

\begin{lemma} \label{zero-1a}
It holds $Q(y)=0$ if and only if $q(\sigma)=0$ for some edge
$\sigma$ of unit length located on $y$ (this edge satisfies
$\sigma_2\sqsubseteq y$).
\end{lemma}
{\it Proof.} The necessary condition is obvious. To prove that this
condition is sufficient, let us suppose that $q(y^n,y^{n+1})=0$
for some $n<l(y)$ but $Q(y)\not=0$. Then by definition
$d(y^n)=1$. Since $Q(y)\not=0$ an extra edge
$(x,z)\in G$ exists such that $x\sqsubseteq y^n$ and $y^{n+1}\sqsubseteq z$.
But, by the construction, this extra edge can not be mounted
to the network $q^{l(z)-1}$ since $d(z^n)=1$.
This contradiction proves the lemma.
$\Box$

For any semimeasure $P$ define
$E_P=\{\omega\in\Omega:\forall n(P(\omega^n)\not=0)\}$ -
the support set of $P$.
It is easy to see that $\bar P(E_P)=\bar P(\Omega)$. By Lemma~\ref{zero-1a}
$E_Q=\Omega\setminus\cup_{d(x)=1}\Gamma_x.$
\begin{lemma} \label{nontriv-1a}
It holds $\bar Q(E_Q)>1-\frac{1}{2}\epsilon$.
\end{lemma}
{\it Proof.} We bound $\bar Q(\Omega)$ from below. Let $R$ be defined by
(\ref{net-base-2}). By definition of the network flow delay function, we have
\begin{eqnarray}
\sum\limits_{u:l(u)=n+1}R(u)=\sum\limits_{u:l(u)=n}(1-d(u))R(u)+
\sum\limits_{\sigma:\sigma\in G,l(\sigma_2)=n+1}q(\sigma)R(\sigma_1).
\label{RR-2a}
\end{eqnarray}
Define an auxiliary sequence
$
S_n=\sum\limits_{u:l(u)=n}R(u)-
\sum\limits_{\sigma:\sigma\in G,l(\sigma_2)=n}q(\sigma)R(\sigma_1).
$
At first, we consider the case $w(p(n),q^{n-1})<n$.
If there is no edge $\sigma\in G$ such that $l(\sigma_2)=n$
then $S_{n+1}\ge S_n$. Suppose that some such edge exists. Define
$$
P(u,\sigma)\Longleftrightarrow l(u)=l(\sigma_2)\&
\sigma_1\sqsubseteq u\&u\not =\sigma_2\&\sigma\in G.
$$
By definition of the network flow delay function, we have
\begin{eqnarray}
\sum\limits_{u:l(u)=n}d(u)R(u)=
\sum\limits_{\sigma:\sigma\in G,l(\sigma_2)=n}d(\sigma_2)
\sum\limits_{u:P(u,\sigma)}R(u)=
\nonumber
\\
=\sum\limits_{\sigma:\sigma\in G,l(\sigma_2)=n}
\frac{d(\sigma_1)}{1-d(\sigma_1)}
\sum\limits_{u:P(u,\sigma)}R(u)\le
\sum\limits_{\sigma:\sigma\in G,l(\sigma_2)=n}d(\sigma_1)R(\sigma_1)=
\nonumber
\\
=\sum\limits_{\sigma:\sigma\in G,l(\sigma_2)=n}q(\sigma)R(\sigma_1).
\label{eqq-1}
\end{eqnarray}
Here we used the inequality
$
\sum\limits_{u:P(u,\sigma)}R(u)\le R(\sigma_1)-
d(\sigma_1)R(\sigma_1)
$
for all $\sigma\in G$ such that $l(\sigma_2)=n$.
Combining this bound with (\ref{RR-2a}) we obtain $S_{n+1}\ge S_n$.

Let us consider the case $w(p(n),q^{n-1})=n$. Then
$
\sum\limits_{u:l(u)=n}d(u)R(u)\le \rho(n)=(n+n_0)^{-2}.
$
Combining (\ref{RR-2a}) and (\ref{eqq-1}) we obtain
$
S_{n+1}\ge S_n-(n+n_0)^{-2}
$
for all $n$. Since $S_0=1$, this implies
$
S_n\ge 1-\sum\limits_{i=1}^{\infty}(i+n_0)^{-2}\ge 1-\frac{1}{2}\epsilon
$
for some sufficiently large constant $n_0$. Since $Q\ge R$, it holds
$$
\bar Q(\Omega)=\inf\limits_{n}\sum\limits_{l(u)=n} Q(u)\ge
\inf\limits_n S_n\ge 1-\frac{1}{2}\epsilon.
$$
Lemma is proved. $\Box$

\begin{lemma} \label{nontriv-1b}
There exists a set $U$ of infinite binary sequences such that
\mbox{$\bar Q(U)\le\epsilon/2$} and for any sequence $\omega\in E_Q\setminus U$ for
each partial computable forecasting system the condition
(\ref{main-1_2}) holds.
\end{lemma}
{\it Proof.} Let $\omega$ be an infinite sequence and let $f$ be a
partial computable forecasting system such that
the corresponding $\phi_t(\omega^{n-1})$ is defined for all $n$. Let
$i={\langle t,s\rangle}$ be a program for computing the rational
approximation $\phi_{t,\kappa_s}$ from below up to $\kappa_s=1/s$.

If $d(\omega^m)=1$ for some $m$ such that $p(m)=i$ then for every
$\beta$ of length
$(1+\lceil (2+\log^{-1}(i+1)\rceil i)m$ such that $\omega^m\sqsubseteq\beta$
there are $<im$ bits hardly predictable by the forecasting program $i$.

We show that $\bar Q$-measure of all intervals generated by
such $\beta$ becomes arbitrary small for
all sufficiently large $i$. Since there are no extra edges $\sigma$ such that
$\omega^m\sqsubseteq\sigma_1$, the measure $\bar Q$ when restricted on interval
$\Gamma_{\omega^m}$
is proportional to the uniform measure. Then by Lemma~\ref{num-hard},
where $\mu=\log^{-1}(i+1)$, $\bar Q$-measure of all such $\beta$ decreases
exponentially by $im$.
Therefore, for each
$j$ there exists a number $m_j$ such that $\bar Q(U_j)\le
2^{-(j+1)}$, where $U_j$ is the union of all intervals
$\Gamma_\beta$ defined by all $\beta$ of length
$(1+\lceil (2+\log^{-1}(i+1))i\rceil)m$ for $m\ge m_j$ containing
$<im$ bits hardly predictable by the forecasting program $i=p(m)$.
Define $U=\cup_{j>k}U_j$, where $k=\lceil-\log_2\epsilon-1\rceil$.
We have $\bar Q(U)<\epsilon/2$.

Define a selection rule $\gamma$ as follows:
\begin{itemize}
\item{}
define $\gamma(\omega^{j-1})=1$ if
$\sigma_1\sqsubseteq\omega^{j-1}\sqsubseteq\sigma_2$ for some
$\sigma\in G(i)$ and the $j$th bit of $\sigma_2$ is hardly predictable
by the forecasting program $i$;
\item{}
define $\gamma(\omega^{j-1})=0$ otherwise.
\end{itemize}

We also define two selection rules
$J_\nu$, where $\nu=0,1$,
\[
  J_\nu(\omega^{j-1})=
  \left\{
    \begin{array}{l}
      1-\nu \mbox{ if } \phi_{t,\kappa_s}(\omega^{j-1})<\frac{1}{2}
    \\
      \nu\mbox{ if } \phi_{t,\kappa_s}(\omega^{j-1})\ge \frac{1}{2}
    \end{array}
  \right.
\]

Suppose that $\omega\not\in U$ and $\phi_t(\omega^n)$ is defined
for all $n$. Then $\omega$ is an $i$-extension of $\omega^n$ for each $n$.
Since for each $n$ the sequence $\omega^n$ is not $i$-closed,
by Lemma~\ref{exten-1} there exists an extra
edge $\sigma\in G(i)$ such that $\sigma_2\sqsubseteq\omega$.
In the following, let $m=l(\sigma_1)$,
$n=(1+\lceil (2+\log^{-1}(i+1))i\rceil )m$.

Then by the construction the selection rule $\delta_\nu(\omega^{j-1})=
\gamma(\omega^{j-1})J_\nu(\omega^{j-1})$, for
$\nu=0$ or for $\nu=1$, selects from a fragment of $\omega$ of length $n$
a subsequence $\omega_{t_1},\dots,\omega_{t_l}$ of length $l\ge im/2$.
Since by definition these bits are hardly predictable, we have
$\omega_{t_j}=1$ for all $j$ such that $1\le j\le l$ if $\nu=0$,
and $\omega_{t_j}=0$ for all these $j$ if $\nu=1$.

Let $\tilde p_j=f(\omega^{j-1})$, $j=1,2,\dots$, be an arbitrary computable
randomizing forecasting system (it is a random variable)
defined on all initial fragments of $\omega=\omega_1\omega_2\dots$.
Then $\phi(\omega^{j-1})=Pr\{\tilde p_j\ge\frac{1}{2}\}$ is a
computable real function. By definition $\phi=\phi_t$ for infinitely
many $t$ and
\begin{eqnarray}\label{ineqq-1}
\phi_{t,\kappa_s}(\omega^{j-1})\le\phi_t(\omega^{j-1})\le
\phi_{t,\kappa_s}(\omega^{j-1})+\kappa_s.
\end{eqnarray}
for all $s$ and $j$. Consider two random variables,
for $\nu=0$ and for $\nu=1$,
\begin{eqnarray*}
\vartheta_{n,\nu}=\sum\limits_{j=1}^n
\delta_\nu (\omega^{j-1}) I_\nu(\tilde p_j)(\omega_j-\tilde p_j).
\label{var-1}
\end{eqnarray*}
Suppose that $l\ge im/2$ holds for $\nu=0$. Then using (\ref{ineqq-1})
we obtain
\begin{eqnarray}
E(\vartheta_{n,0})\ge\sum\limits_{j=m+1}^n
\delta_0(\omega^{j-1})Pr\{\tilde p_j<\frac{1}{2}\}\frac{1}{2}-m\ge
\nonumber
\\
\ge\frac{im}{4}(\frac{1}{2}-\kappa_s)-m
\label{avv-1}
\end{eqnarray}
Since $n=(1+\lceil (2+\log^{-1}(i+1))i\rceil )m$, $i$ can be arbitrary large
and we visit any pair $i=\langle t,s\rangle$ infinitely often,
we obtain from (\ref{avv-1})
\begin{eqnarray}
\limsup\limits_{n\to\infty}\frac{1}{n}E(\vartheta_{n,0})\ge 1/16.
\label{iin-1}
\end{eqnarray}
Analogously, if $\nu=1$ we obtain
\begin{eqnarray}
\liminf\limits_{n\to\infty}\frac{1}{n}E(\vartheta_{n,1})\le -1/16.
\label{iin-2}
\end{eqnarray}

The martingale strong law of large numbers says that for $\nu=0,1$
with $Pr$-probability one
\begin{eqnarray}
\frac{1}{n}\sum\limits_{j=1}^n\delta_\nu(\omega^{j-1})
I_\nu(\tilde p_j)(\omega_j-\tilde p_j)-\frac{1}{n}E(\vartheta_{n,\nu})\to 0
\label{mart-1}
\end{eqnarray}
as $n\to\infty$.
Combining (\ref{iin-1}), (\ref{iin-2}) and (\ref{mart-1}) we obtain
(\ref{main-1_2}).

Lemma~\ref{nontriv-1b} and Theorem~\ref{main-1} are proved.
$\Box$

The following theorem is a generalization of the result from
V'yugin~\cite{Vyu98} for partial defined computable deterministic
forecasting systems.
\begin{theorem}\label{main-2}
For any $\epsilon>0$ a probabilistic algorithm $(L,F)$ can be
constructed, which with probability $\ge 1-\epsilon$ outputs an
infinite binary sequence $\omega=\omega_1\omega_2\dots$ such that
for every partial deterministic forecasting algorithm $f$ defined
on all initial fragments of the sequence $\omega$ a computable
outcome-based selection rule $\delta$ exists defined on all these
fragments such that
\begin{eqnarray}
\limsup_{n\to\infty}\left|\frac{1}{n}
\sum_{i=1}^{n}\delta(\omega^{i-1})(\omega_i-f(\omega^{i-1}))\right|\ge 1/8.
\label{cal-2}
\end{eqnarray}
\end{theorem}
The proof of this theorem is based on the same construction.

\section{Acknowledgements}

Author thanks an anonymous referee pointing out the connection of
the results of this paper with works~\cite{Lehr97}
and~\cite{sansmovo03}.

This research was partially supported by Russian foundation
for fundamental research: 06-01-00122-a.

\end{document}